\documentclass[letterpaper]{article} 
\usepackage{aaai24}  
\usepackage{times}  
\usepackage{helvet}  
\usepackage{courier}  
\usepackage[hyphens]{url}  
\usepackage{graphicx} 
\usepackage{natbib}  
\usepackage{caption} 
\usepackage{algorithm}
\usepackage{algorithmic}
\usepackage{amsmath}
\usepackage{booktabs}
\usepackage{amsfonts}
\usepackage{array}
\usepackage{multirow}
\usepackage{subcaption}
\usepackage{enumitem}
\usepackage{arydshln}
\usepackage{textpos} 
\usepackage{newfloat}
\usepackage{listings}
\DeclareCaptionStyle{ruled}{labelfont=normalfont,labelsep=colon,strut=off} 
\floatstyle{ruled}
\newfloat{listing}{tb}{lst}{}
\floatname{listing}{Listing}
\title{An Empirical Study of CLIP for Text-based Person Search}
\author{
    Min Cao\textsuperscript{\rm 1},
    Yang Bai\textsuperscript{\rm 1},
    Ziyin Zeng\textsuperscript{\rm 1},
    Mang Ye\textsuperscript{\rm 2},
    Min Zhang\textsuperscript{\rm 1}
}
\affiliations{
    \textsuperscript{\rm 1}School of Computer Science and Technology, Soochow University\\
    \textsuperscript{\rm 2}School of Computer Science, Wuhan University
}

\usepackage{bibentry}

\begin{document}

\maketitle

\begin{abstract}
Text-based Person Search (TBPS) aims to retrieve the person images using natural language descriptions. 
Recently, Contrastive Language Image Pretraining (CLIP), a universal large cross-modal vision-language pre-training model, has remarkably performed over various cross-modal downstream tasks due to its powerful cross-modal semantic learning capacity. TPBS, as a fine-grained cross-modal retrieval task, is also facing the rise of research on the CLIP-based TBPS. 
In order to explore the potential of the visual-language pre-training model for downstream TBPS tasks, this paper makes the first attempt to conduct a comprehensive empirical study of CLIP for TBPS and thus contribute a straightforward, incremental, yet strong TBPS-CLIP baseline to the TBPS community.
We revisit critical design considerations under CLIP, including data augmentation and loss function.
The model, with the aforementioned designs and practical training tricks, can attain satisfactory performance \textbf{without any sophisticated modules}. Also, we conduct the probing experiments of TBPS-CLIP in model generalization and model compression, demonstrating the effectiveness of TBPS-CLIP from various aspects.
This work is expected to provide empirical insights and highlight future CLIP-based TBPS research.

\end{abstract}

\begin{textblock*}{.8\textwidth}[.5,0](0.5\textwidth, -.560\textwidth)
\centering
{\small Code: \url{https://github.com/Flame-Chasers/TBPS-CLIP}}
\end{textblock*}

\section{Introduction}

Text-based Person Search (TBPS) retrieves the corresponding person images from a large-scale image database given a textual description. 
The task is gradually gaining extensive attention~\cite{jiang2023cross,bai2023rasa} due to its potential applications in searching for suspects, locating lost children, \emph{etc}. 
As a fine-grained retrieval task, TBPS requires the ability to achieve effective cross-modal alignment and efficient cross-modal retrieval for practical applications, both of which bring challenges.

For replying to these challenges, many methods~\cite{zhang2018deep, ding2021semantically, zhao2021weakly} focus on projecting representations extracted from each modality into one shared space.
However, most of them only utilize unimodal pre-trained models as the backbones (\emph{e.g.}, LSTM or BERT for the text encoder, Resnet-50 or ViT for the image encoder), and ignore the powerful Vision-Language Pre-training (VLP) models equipped with an adequate understanding of cross-modal alignment.
In recent years, VLP methods~\cite{lu2019vilbert, li2022blip} have demonstrated remarkable performance on various cross-modal downstream tasks, including visual question answering, image captioning, image-text retrieval, \emph{etc}. Consequently, it has become the dominant paradigm for solving these tasks. 
Considering that the research under VLP for TBPS~\cite{bai2023text,bai2023rasa} is now in its infancy, this paper aims to fully exploit the potential of VLP for TBPS.

\begin{figure}[t]
\setlength{\abovecaptionskip}{0.05cm}
\centerline{\includegraphics[width=0.75\linewidth,height=0.8\linewidth]{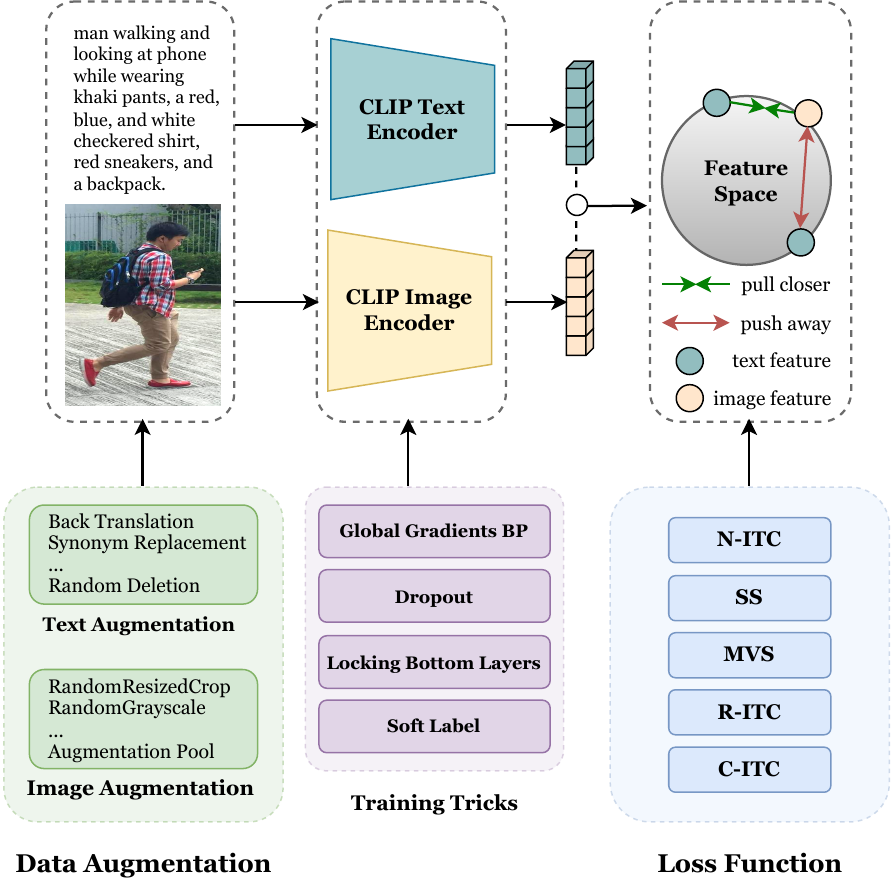}}
\caption{Overview of the empirical study on CLIP.} \label{fig:head}
\vspace{-0.5cm}
\end{figure}

Among VLP methods, CLIP~\cite{radford2021learning} is one of the most prominent representatives and has demonstrated impressive performance on various downstream tasks~\cite{wang2022cris, wang2022clip}.
In addition, CLIP has the benefit of efficient retrieval due to its independent encoding for each modality. 
These inspire some researchers to explore CLIP-based TBPS methods~\cite{wang2023exploiting, jiang2023cross}.
They specialize in incorporating sophisticated modules into CLIP to enhance performance, lacking the full exploitation of CLIP's pre-trained knowledge and the full utilization of its powerful cross-modal semantic learning capacity.
This paper returns to the origin of CLIP and probes the potential of CLIP tuning for TBPS. 
For this, we conduct a comprehensive empirical study of CLIP for TBPS from two aspects: data augmentation and loss function. 

1) Data augmentation.
It is a prevalent and effective technique to increase the generalization of models by learning augmentation-invariant representations between the original input and the augmented version. 
Until now, none of the TBPS methods have deeply and systematically explored this technique. Instead, they typically employ a simple utilization of random horizontal flip for the image while not applying any augmentation to the text~\cite{wang2020vitaa, chen2022tipcb, ding2021semantically}.
In this study, we comprehensively evaluate various data augmentation strategies, thereby deriving a powerful data augmentation strategy for TBPS.

2) Loss function.
Designing rational and practical loss functions is critical to improving performance and has been an increasingly active research direction in TBPS community~\cite{zhang2018deep, bai2023rasa}.
We take CLIP as a hotbed and conduct a series of probing studies to analyze the effectiveness of various loss functions in TBPS.
Unlike the loss functions in existing TBPS methods that are well-designed mainly from exploring the TBPS task and belong to the task-oriented loss functions, the loss functions probed in this study are primarily inspired by VLP communities and are pretty generic to various cross-modal tasks.

These empirical studies above, combined with other valuable tricks detailed in later sections, enable us to develop a strong TBPS-CLIP baseline.
\textbf{Unlike other methods of designing sophisticated modules, TBPS-CLIP attains competitive performance under a very lightweight and low-cost architecture, in which merely several common training tricks, data augmentations and loss functions are added into CLIP.}
In order to fully demonstrate the effectiveness and generalization of TBPS-CLIP, we further develop some valuable additional probing experiments.

1) Model generalization.
For one thing, CLIP has become a baseline of various tasks due to its effectiveness and simplicity, likewise, TBPS-CLIP is targeted as the baseline of TBPS, for which its effectiveness as the baseline is experimentally proved. 
For another thing, beyond only employing CLIP for the supervised TBPS, we provide a preliminary exploration of the few-shot TBPS under CLIP and the superiority of TBPS-CLIP in the few-shot setting is also proved.

2) Model compression. 
We provide insights into the internal properties of TBPS-CLIP by investigating the contribution of each module to the final retrieval performance. These insights offer the guidance for compression of TBPS-CLIP.


In summary, this paper mainly contributes a strong TBPS-CLIP baseline to the TBPS community.
TBPS-CLIP has features of high-performance, lightweight and low-cost architecture and ease of use.
Besides these, we show its advantage in model generalization and model compression.
All these manifest the applicability of TBPS-CLIP in both the academy and industry.
Lots of experiments are expected to provide empirical insights for future research.


\section{Related Work}
\subsection{Text-based Person Search}

Adopting the unimodal pre-trained models as backbones is a conventional pattern in TBPS methods~\cite{wang2020vitaa, wu2021lapscore, li2017identity,li2022learning}. Most of these methods adopt ResNet-50 or ViT as the image encoder and LSTM or BERT as the text one. On the basis of these unimodal backbones, some works~\cite{zhang2018deep, sarafianos2019adversarial, li2017identity,li2022learning} design specific loss functions for learning discriminative representations.
Meanwhile, some extract fine-grained information from the image and text for aligning cross-modal fine-grained features. Specifically, partitioning images into horizontal stripes is one way to obtain fine-grained image information~\cite{chen2022tipcb, zheng2020dual, ding2021semantically}, whereas a set of noun phrases can be parsed by utilizing NLTK Toolbox to obtain fine-grained text information~\cite{wang2020vitaa}.

Witnessing VLP's great success on cross-modal tasks in recent years, researchers~\cite{han2021text, yan2022clip, wang2023exploiting, jiang2023cross,bai2023rasa} have begun pushing the frontier of TBPS solutions with VLP.
Han \emph{et al}.~\cite{han2021text} propose to employ CLIP~\cite{radford2021learning} as the backbone and appends a Bi-GRU after its text encoder for better text feature encoding;
CFine~\cite{yan2022clip} embraces the image encoder from CLIP to enrich cross-modal correspondence information, and uses the text encoder replaced with BERT to avoid intra-modal information distortion; 
TP-TPS~\cite{wang2023exploiting} probes the textual potential of CLIP in TBPS by aligning the images to the constructed multi-integrity descriptions and attribute prompts;
IRRA~\cite{jiang2023cross} designs an implicit relation reasoning module above CLIP to align fine-grained information across modalities;
RaSa~\cite{bai2023rasa} develops two novel losses under ALBEF~\cite{li2021align} backbone for TBPS;
Bai \emph{et al}.~\cite{bai2023text} leverage VLP knowledge to solve TBPS without parallel image-text data.
In contrast to these methods, which concentrate on exploiting specific modules beyond VLP to improve TBPS performance, this paper aims to fully mine the potential of CLIP itself through an empirical study of CLIP for TBPS, resulting in TBPS-CLIP with competitive performance without introducing any sophisticated modules into CLIP.

\subsection{Vision-Language Pre-training}
In recent years, VLP~\cite{lu2019vilbert, li2022blip} has emerged as a dominant solution for a wide range of cross-modal tasks, including but not limited to image captioning, visual question answering, and visual grounding. VLP methods capitalize on the power of large-scale pre-training with massive amounts of image-text pairs, which enables the models to learn reliable cross-modal representations that can further be fine-tuned for various downstream tasks.

Among various VLP methods, the Contrastive Language-Image Pre-training (CLIP)~\cite{radford2021learning} stands out for its excellent performance on various cross-modal tasks~\cite{wang2022cris, wang2022clip}. 
Furthermore, several works are carried out to enhance the data efficiency of CLIP. 
For example, SLIP~\cite{mu2022slip} introduces a self-supervised learning loss alongside the contrastive loss in CLIP; FILIP~\cite{yao2021filip} employs a token-wise maximum similarity between visual and textual tokens to guide the contrastive loss in CLIP; DeCLIP~\cite{li2021supervision} exploits multiple supervisions, including self-supervision, multi-view supervision and nearest-neighbor supervision, to replace the single contrastive supervision.
In addition, some works concentrate on exploring CLIP's few-shot capabilities.
For example, CoOp~\cite{zhou2022learning} utilizes learnable vectors as the prompts and achieves significant performance gain in a few-shot regime, and CLIP-Adapter~\cite{gao2021clip} appends the learnable module to CLIP and achieves decent few-shot results by only fine-tuning this module.



\section{Empirical Studies}
\label{sec:3}
Grounded on CLIP, we first introduce some practical training tricks to strengthen the CLIP baseline in Section~\ref{sec:3.1}, and then elaborate data augmentations and loss functions for the empirical study in Section~\ref{sec:3.2} and Section~\ref{sec:3.3}, respectively. 
We finally study the model generalization and model compression in Section~\ref{sec:3.5} and Section~\ref{sec:3.4}, respectively. 
The overview of the model is illustrated in Figure~\ref{fig:head}.

\subsection{Training Tricks} \label{sec:3.1}
We investigate four common training tricks around CLIP.
They are: \emph{global gradients back-propagation}, \emph{dropout}, \emph{locking bottom layers} and \emph{soft label}.
\emph{These tricks are detailed in the Appendix.}

\subsection{Data Augmentations} \label{sec:3.2}



\paragraph{Image Augmentation.}

We classify image augmentations into two groups: removal and alteration. 
The first group has operations that remove some information from the image, including \emph{RandomResizedCrop}, \emph{RandomErasing}, \emph{RandomGrayscale} and \emph{GaussianBlur}.
The second alters the color or orientation of the image with keeping the main content, including \emph{ColorJitter}, \emph{RandomHorizontalFlip}, \emph{RandomVerticalFlip} and \emph{RandomRotation}.
\emph{These common image augmentations are detailed in the Appendix.}



In addition, it should be noted that the simultaneous use of multiple image augmentations may bring heavy distortion to the original image and hurt performance.
Given that, we consider another series of augmentations, they are:
\begin{itemize}
    \item \textit{AutoAugment~\cite{cubuk2018autoaugment}} automatically search for the best augmentation policy, for which reinforcement learning is applied in some augmentation policies. We use its default settings on PyTorch in the experiment.
    
    \item \textit{RandAugment~\cite{cubuk2020randaugment}} removes the searching stage in AutoAugment and randomly selects from a pool of augmentation operations. The 2 hyperparameters are the number of augmentations $N$ applied on each image and the magnitude $M$ in the range [1, 31]. We follow the default settings on PyTorch, setting $N$ to 2 and $M$ to 9.
    \item \textit{TrivialAugment~\cite{muller2021trivialaugment}} further drops the requirement of setting hyperparameters in RandAugment, and instead randomly selects one augmentation and its magnitude to apply on each image. Therefore, it is completely parameter-free.
    \item
    \textit{An augmentation pool strategy} is designed in this paper, inspired by the above-mentioned automatic data augmentations. Specifically, two augmentations are randomly selected and applied to each image.
\end{itemize}

\paragraph{Text Augmentation.}
In contrast to image augmentation, there is a smaller pool of available text augmentations due to the abstract and discrete nature of language. 

\begin{itemize}
    \item \textit{Back translation} translates the original text to a specific language and then translates it back, by which we can obtain more diverse textual descriptions while preserving its original meaning.
    In our experimental setup, we adopt French as the intermediate language.
    It has a relatively closer form to English and introduces fewer changes to the translated back text in semantics than others.
    
    \item \textit{Synonym replacement} randomly chooses some words from the sentence, and then replaces each of them with a synonym chosen at random.

    \item \textit{Random insertion} randomly chooses some words from the sentence and the synonyms of selected words are then inserted into random positions of the sentence.
    
    \item \textit{Random swap} selects two words from the sentence at random and interchanges their positions. 

    \item \textit{Random deletion} randomly removes each word in a sentence.

    \item \textit{EDA}~\cite{wei2019eda} randomly selects one from synonym replacement, random insertion, random swap and random deletion and applies it to the sentence.
\end{itemize}

\subsection{Loss Functions} \label{sec:3.3}

CLIP equips with an image-text contrastive loss to pull positive samples together
while pushing negative ones apart.
Further, we normalize the label in the loss and obtain the normalized image-text contrastive loss (N-ITC):
\begin{small}
\begin{equation}
    \mathcal{L}_{N-ITC} = -\frac{1}{2N}( \sum_{i=1}^{N}\sum_{j=1}^{N} \hat{q}_{i,j} \log{p_{i,j}} + \sum_{i=1}^{N}\sum_{j=1}^{N} \hat{q}_{j,i} \log{p_{j,i}}),
    \label{eq:itc}
\end{equation}
\end{small}
where $\hat{q}_{i,j}$ is normalized by ${q_{i,j}}/{\sum_{k=1}^{N} q_{i,k}}$ and $q_{i,j}$ is the ground-truth label ($1$ for positive pair and $0$ for negative one).
$N$ is the number of samples, and 
$p_{i,j}$ represents the pseudo-label that is the probability of matching the image $I_i$ to the text $T_j$ and the reverse applies in $p_{j,i}$,
\begin{small}
\begin{equation}
    p_{i,j} = \frac {\exp( f_{I_{i}} \cdot f_{T_{j}} /\tau)}
        {\sum_{k=1}^{N} \exp( f_{I_{i}} \cdot f_{T_{k}} /\tau)}, 
    p_{j,i} = \frac {\exp( f_{T_{i}} \cdot f_{I_{j}} /\tau)}
        {\sum_{k=1}^{N} \exp( f_{T_{i}} \cdot f_{I_{k}} /\tau)},
\end{equation}
\end{small}
where $f_*$ is the $\ell_2$-normalized representations of the sample, $\tau$ is the learnable temperature parameter.

Apart from the N-ITC, we study other losses in two directions.
One focuses on enhancing data efficiency,
and the other targets optimizing the relationship between samples for improving performance.

\subsubsection{Improving Data Efficiency}
\paragraph{Self-Supervision.}
The self-supervised loss (SS) aims at maximizing the similarity between two different augmentations of an image and prompts learning robust feature representations with limited image data.
It has demonstrated effectiveness in various visual tasks~\cite{chen2020simple, chen2021exploring}, and motivates us to explore it in TBPS. 
Specifically, the SS is defined as,
\begin{equation}
\mathcal{L}_{SS} = -\frac{1}{2N} \sum_{i=1}^{2N} \log \frac{\exp(sim(z_i, z_j) / \tau_s)}{\sum_{k=1, k\neq i}^{2N} \exp(sim(z_i, z_k) / \tau_s)},
\end{equation}
where $\tau_s$ is a hyper-parameter and set to $0.1$, and $z_i$ and $z_j$ are the feature representations of two augmentations of the sample. 
The SS can be applied on the image or text or both of them, denoted as SS-I, SS-T and SS-IT, respectively.

\paragraph{Multi-View Supervision.}
The N-ITC in Equation~\ref{eq:itc} only leverages one augmented view of data. 
Inspired by DeCLIP~\cite{li2021supervision}, more views of samples can provide extra supervision to motivate the potential of limited data.
Specifically, let $\tilde{I}$ and $\tilde{T}$ denote another differently augmented view of $I$ and $T$ by the data augmentation technology, the N-ITC can be applied on ($\tilde{I}, T$) or ($I, \tilde{T}$) or ($\tilde{I}, \tilde{T}$), denoted as the multi-view supervision loss of image (MVS-I), multi-view supervision loss of text (MVS-T), multi-view supervision loss of image and text (MVS-IT), respectively.

\subsubsection{Optimization for Retrieval}

\paragraph{Reversed Image-Text Contrastive Loss.}

The N-ITC in Equation~\ref{eq:itc} is roughly equivalent to optimizing $D_{\text{KL}}(Q\| P)$, where $Q$ represents the label distribution and $P$ denotes the optimized similarity distribution. 
It mainly focuses on assigning the image-text pair to high similarity probability when its label distribution places high probability (\emph{i.e.}, positive pair).
As a supplement to it, we take inspiration from CMPM~\cite{zhang2018deep} and optimize $D_{\text{KL}}(P\| Q)$, which have $P$ and $Q$ reversed in the N-ITC, and the reversed image-text contrastive loss (R-ITC) considers separating the negative pairs.
Specifically, the R-ITC is formulated as:
\begin{equation}
\resizebox{0.98\columnwidth}{!}{%
$\begin{aligned}
    \mathcal{L}_{R-ITC} &= \frac{1}{2} (D_{\text{KL}}(p_{i,j} \| \hat{q}_{i,j})+ D_{\text{KL}}(p_{j,i} \| \hat{q}_{j,i})) \\
    &= \frac{1}{2N} (\sum_{i=1}^{N}\sum_{j=1}^{N} p_{i,j} \log{\frac{p_{i,j}}{\hat{q}_{i,j}+\epsilon}}+ \sum_{i=1}^{N}\sum_{j=1}^{N} p_{j,i} \log{\frac{p_{j,i}}{\hat{q}_{j,i}+\epsilon}}),
\end{aligned}$%
}
\end{equation}
where $\epsilon$ is a small number to avoid having a zero in the denominator.

\begin{figure}
\setlength{\abovecaptionskip}{0.05cm}
\centerline{\includegraphics[width=0.75\columnwidth]{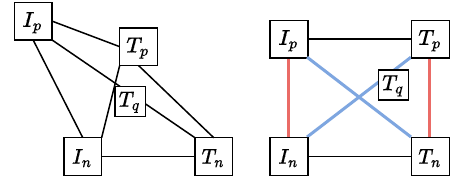}}
\caption{Difference between C-ITC (right-hand) and other matching losses (left-hand), \emph{e.g.}, N-ITC and R-ITC. $(I_p, T_p)$ and $(I_n, T_n)$ are two paired cross-modal data and the query text $T_q$ shares the same identity with $(I_p, T_p)$.}
\label{fig:cyclip}
\vspace{-0.5cm}
\end{figure}

\paragraph{Cyclic Image-Text Contrastive Loss.}
The general matching loss (\emph{e.g.}, N-ITC and R-ITC) focuses on optimizing the relationship between the image and text, which may cause the image and text to be irregularly positioned in the representation space, and carries the risk that the query text $T_q$ mistakenly retrieves the negative image $I_n$, while it should match the positive image $I_p$ due to its high similarity with the text $T_p$, as illustrated on the left-hand side of Figure~\ref{fig:cyclip}. 
Following CyCLIP~\cite{goel2022cyclip}, which enhances geometrical consistency in data representation space, we study the cyclic image-text contrastive loss (C-ITC) to mitigate the above problem.
The right-hand side of Figure~\ref{fig:cyclip} shows the geometry of the resulting representation space.


Specifically, the learned representations are regularized in two ways: in-modality and cross-modality. 
For the in-modality regularization, the loss enables reducing the gap between the distance of the similarity of two images and that between the corresponding texts:
\begin{equation}
    \mathcal{L}_{C^I-ITC}
    = \frac{1}{N} \sum_{i=1}^N \sum_{j=1}^N (sim(I_i, I_j) - sim(T_i, T_j))^2.
\end{equation}

For the cross-modality regularization, given two image-text pairs, the gap between their distances is minimized:
\begin{equation}
    \mathcal{L}_{C^C-ITC}
    = \frac{1}{N} \sum_{i=1}^N \sum_{j=1}^N (sim(I_i, T_j) - sim(I_j, T_i))^2.
\end{equation}

Together, the C-ITC can be formulated as:
\begin{equation}
    \mathcal{L}_{C-ITC} = \mathcal{L}_{C^I-ITC} + \mathcal{L}_{C^C-ITC}.
\end{equation}




\subsection{Model Generalization} \label{sec:3.5}
Apart from the empirical study on data augmentation and loss function, 
we also demonstrate the model generalization from two sides.
Specifically, we first apply the proposed TBPS-CLIP to other TBPS methods for verifying the generalization of TBPS-CLIP as the baseline, and then fine-tune TBPS-CLIP based on small amounts of TBPS training data for verifying the generalization of TBPS-CLIP in the few-shot setting. 
We notice that several CLIP variants have been proposed to enhance CLIP's few-shot capability.
Hence, we also conduct the empirical study on two representative few-shot CLIP methods, CoOp~\cite{zhou2022learning} and CLIP-Adapter~\cite{gao2021clip}, for the performance comparison.
\emph{The overview of the two methods is shown in the Appendix.}


\subsection{Model Compression}  \label{sec:3.4}

We provide insight into the model by evaluating the role of each module to the final performance, which is valuable for model compression in real-world applications.
Two evaluation metrics~\cite{wang2020rethinking} are adopted.
The first one evaluates a specific module's contribution by removing the module and observing the performance drop. 
The second one examines the module's importance by measuring how closely the module's weights can approach their initial values while maintaining a certain level of performance. 
\emph{We elaborate on the two metrics in the Appendix.}

\section{Experiments}
The experimental analyses of the empirical studies are detailed in this section. 
Also, although our studies are directed at common technologies to keep the model simple, we provide insights from the TBPS-specific view in this section, \emph{i.e.,} discussing why these technologies can work in TBPS.

Comparisons with other methods are carried out on three benchmark datasets, \emph{i.e.}, CUHK-PEDES~\cite{li2017person}, ICFG-PEDES~\cite{ding2021semantically} and RSTPReid~\cite{zhu2021dssl}, while extensive ablation studies are on the most common CUHK-PEDES.
We use Rank-$k$ and mean Average Precision (mAP) to evaluate the performance.
\emph{The further introduction of dataset, evaluation metric and implementation detail are shown in the Appendix.}





\begin{table}[t]
\setlength{\abovecaptionskip}{-0.01cm}
\begin{center}
{\caption{Ablations of training tricks on CUHK-PEDES. CLIP$^*$ represents CLIP with all four training trick.}\label{table:tricks}}
\resizebox{0.85\columnwidth}{!}{
\begin{tabular}{l!{\vrule}cccc}
\toprule
Methods & Rank-1 & Rank-5 & Rank-10 & mAP   \\
\midrule
CLIP & 60.67 & 81.99 & 88.87 & 54.72 \\ 
CLIP + GlobalGrad & 63.66 &  \textbf{84.08} & 90.11 & 56.46 \\
CLIP + Dropout & 61.06 & 82.00 & 88.95 & 55.10 \\
CLIP + LockBL & 61.27 & 82.23 & 88.79 & 55.43 \\
CLIP + SLabel & 61.53 & 81.99 & 88.71 & 55.22 \\
\midrule
CLIP$^*$ &  \textbf{64.34} & 84.05 &  \textbf{90.51} &  \textbf{57.53} \\
\bottomrule
\end{tabular}
}
\end{center}
\vspace{-0.5cm}
\end{table}

\subsection{Ablations of Training Tricks} \label{sec:4.3}

Starting from CLIP, we first perform experimental investigations on the training tricks discussed in Section~\ref{sec:3.1}. 
As shown in Table~\ref{table:tricks}, all training tricks introduce a positive impact into CLIP on the performance.

\subsection{Ablations of Data Augmentation}\label{sec:exp-aug}

We show the optimal results of each augmentation, and analyze their effects.
\emph{The ablation study of the augmentations with different hyperparameters is shown in the Appendix.}

\paragraph{Image Augmentations.} 
Table~\ref{table:image_aug} studies the effect of image augmentations.
(1) In the removal augmentations, RandomResizedCrop and RandomErasing all gain performance enhancement. 
They randomly crop/erase the part of the image, as a result, the augmented image highlights the local details and indirectly encourages the cross-modal fine-grained learning of TBPS.
Unexpectedly, RandomGrayscale, eliminating all colour information from the image, also leads to improved results.
The augmented image without colour by RandomGrayscale is inputted into the model, and the model has to focus on other information (such as texture and shape) during training.
It is well-known that colour information plays a critical role in TBPS~\cite{wu2021lapscore,wang2022caibc}, yet we can learn from the empirical results that other information, except for colour, is also valuable for person retrieval.
In addition, GaussianBlur, smoothing the fine-grained details of the image, significantly impairs performance.
The smoothing operated on the whole image brings the loss of fine-grained information, which is crucial for TBPS.
(2) In the alternation augmentations, most are beneficial for the performance, including ColorJitter-BCS, RandomHorizontalFlip and RandomRotation. 
They enrich the image diversity without altering the image semantics and strengthen the model's robustness against various images, thus resulting in better performance. By contrast, ColorJitter-Hue and RandomVerticalFlip hurt performance. Both of them change the image semantics. 
The former alters the color, making the color information in the image inconsistent with the corresponding textual description, while the latter dramatically changes the rough shape of the image.
(3) Beyond applying the single augmentation to the image, we conduct experiments of applying multiple augmentations.
Intuitively, we can stack the above effective augmentations to the image (Stacking Together), which leads to performance improvement.
In addition, we can adapt automatic image augmentation strategies AutoAugment~\cite{cubuk2018autoaugment}, RandAugment~\cite{cubuk2020randaugment} and TrivialAugment~\cite{muller2021trivialaugment}.
Nevertheless, the proposed augmentation pool that randomly selects $2$ effective augmentations each time gets the best results.

\begin{table}
\setlength{\abovecaptionskip}{-0.01cm}
\begin{center}
{\caption{Ablations of image augmentations on CUHK-PEDES. `Stacking Together' applies multiple removal and alternation augmentations with the `$\checkmark$' symbol into the image, while `Augmentation Pool' randomly selects two of them in each iteration. `AutoAug', `RandAug' and `TrivialAug' are applied at the default setting.}\label{table:image_aug}}
\resizebox{0.99\columnwidth}{!}{
\begin{tabular}{l!{\vrule}cccc}
\toprule
Methods  & Rank-1 & Rank-5 & Rank-10 & mAP \\
\midrule
CLIP$^*$ & 64.34 &  84.05 &  90.51 &  57.53 \\
\midrule
\multicolumn{5}{l}{\textit{The group of removal:}} \\
\midrule
RandomResizedCrop (\checkmark) & 65.29 &  85.01 &  90.68 & 58.62 \\
RandomErasing (\checkmark) & 64.64 &  84.34 &  90.74 &  58.08 \\
RandomGrayscale (\checkmark) &  65.29 &  84.67 &  90.56 & 58.18 \\
GaussianBlur & 55.30 & 77.86 & 85.88 & 49.82 \\
\midrule
\multicolumn{5}{l}{\textit{The group of alternation:}} \\
\midrule
ColorJitter-BCS (\checkmark) &  64.86 &  84.65 &  90.47 &  57.85 \\
ColorJitter-Hue & 64.12 & 84.81 & 90.87 & 57.41 \\
\hdashline[1pt/1pt]
RandomHorizontalFlip (\checkmark) & 65.48 &  84.86 & 90.48 & 58.46 \\
RandomVerticalFlip & 64.23 & 83.98 & 90.16 & 57.35 \\
RandomRotation (\checkmark) &  65.50 &  85.20 &  91.13 &  58.92 \\
\midrule
\multicolumn{5}{l}{\textit{Applying multiple augmentations:}} \\
\midrule
Stacking Together & 65.92 & 85.33 & 90.89 &  59.43 \\
AutoAug~\cite{cubuk2018autoaugment} & 65.43 & 84.84 & 91.07 & 58.55 \\
RandAug~\cite{cubuk2020randaugment} & 65.58 &  \textbf{85.72} & 91.00 & 59.08 \\
TrivialAug~\cite{muller2021trivialaugment} & 65.59 & 84.58 & 90.81 & 58.58 \\
 Augmentation Pool &  \textbf{66.13} & 85.38 &  \textbf{91.21} & \textbf{59.30} \\
\bottomrule
\end{tabular}
}
\end{center}
\vspace{-0.1cm}
\end{table}

\begin{table}[t]
\setlength{\abovecaptionskip}{-0.01cm}
\begin{center}
{\caption{Ablations of text augmentations on CUHK-PEDES. `Stacking Together' applies multiple augmentations with the `$\checkmark$' symbol into the text.}\label{table:text_aug}}
\resizebox{0.95\columnwidth}{!}{
\begin{tabular}{l!{\vrule}cccc}
\toprule
Methods & Rank-1 & Rank-5 & Rank-10 & mAP \\
\midrule
CLIP$^*$ & 64.34 &  84.05 &  90.51 &  57.53 \\
\midrule
Back Translation (\checkmark) & 64.77 &  \textbf{84.67} &  90.56 &  57.49 \\
Synonym Replacement & 63.95 & 83.76 & 89.96 & 56.90 \\
Random Insertion & 63.71 & 84.29 & 90.03 & 56.77 \\
Random Swap & 56.41 & 79.08 & 86.14 & 49.29 \\
Random Deletion (\checkmark) &  65.53 &  84.70 &  \textbf{90.87} &  57.90 \\
EDA~\cite{wei2019eda} & 63.43 & 83.77 & 90.27 & 56.15 \\
\midrule
 Stacking Together &  \textbf{65.72} & 84.62 &  90.76 & \textbf{58.68} \\
\bottomrule
\end{tabular}
}
\end{center}
\vspace{-0.5cm}
\end{table}

\begin{table}[t]
\setlength{\abovecaptionskip}{-0.01cm}
\begin{center}
{\caption{Results of combining optimal augmentations within each modality.}\label{table:aug_all}}
\resizebox{0.93\columnwidth}{!}{
\begin{tabular}{l!{\vrule}cccc}
\toprule
Methods & Rank-1 & Rank-5 & Rank-10 & mAP \\
\midrule
 CLIP$^*$ &  64.34 &   84.05 &  90.51 &  57.53 \\
\midrule
Image Aug & 66.13 & 85.38 & 91.21 & 59.30 \\
Text Aug & 65.72 & 84.62 & 90.76 & 58.68 \\
  Image \& Text Aug &  \textbf{66.78} &  \textbf{85.98} &  \textbf{91.23} &  \textbf{59.68} \\
\bottomrule
\end{tabular}
}
\end{center}
\vspace{-0.1cm}
\end{table}

\begin{table}[t]
\setlength{\abovecaptionskip}{-0.01cm}
\begin{center}
{\caption{Ablations of loss functions on CUHK-PEDES. `Stacking Together' denotes that multiple losses with the `$\checkmark$' symbol are used to CLIP$^*$+Aug .}\label{table:loss}}
\resizebox{0.48\textwidth}{!}{
\begin{tabular}{l!{\vrule}cccc}
\toprule
Methods & Rank-1 & Rank-5 & Rank-10 & mAP \\
\midrule
CLIP$^*$+Aug & 66.78 & 85.98 & 91.23 & 59.68 \\
\midrule
N-ITC (\checkmark) & 66.91 & 85.98 & 91.68 & 60.01 \\
\midrule
\multicolumn{5}{l}{\textit{Improving data efficiency:}} \\
\midrule
SS-I (\checkmark) & 67.54 & 86.13 &  \textbf{91.78} & 60.94 \\
SS-T & 65.97 & 85.40 & 90.72 & 59.10 \\
SS-IT & 66.81 & 86.09 & 91.63 & 59.82 \\
\hdashline[1pt/1pt]
MVS-I (\checkmark)\rule{0pt}{8pt} & 67.50 & 86.45 & 91.63 & 60.84 \\
MVS-T & 66.46 & 85.96 & 91.13 & 60.24 \\
MVS-IT & 67.01 & 85.66 & 91.33 & 59.83 \\
\midrule
\multicolumn{5}{l}{\textit{Optimization for retrieval:}} \\
\midrule
R-ITC (\checkmark) &  68.19 & 86.01 & 90.94 & 60.90 \\
C-ITC (\checkmark) & 67.15 & 86.31 & 92.04 & 60.58 \\
\midrule
\multicolumn{5}{l}{\textit{Applying multiple losses:}} \\
\midrule
 Stacking Together (TBPS-CLIP) &  \textbf{69.54} &  \textbf{86.99} & 91.24 &  \textbf{61.57} \\
\bottomrule
\end{tabular}
}
\end{center}
\vspace{-0.5cm}
\end{table}

\paragraph{Text Augmentations.} 
As presented in Table~\ref{table:text_aug}, the effective text augmentations for TBPS are the back translation and random deletion.
The back translation directly enriches the forms of the original texts, and the random deletion plays a role of regularization by randomly dropping some words,
both of which positively impact performance.
Accordingly, applying the two together (Stacking Together) leads to an improvement of $1.38\%$ at Rank-1.
On the contrary, the synonym replacement, random insertion and random swap all hurt performance by a margin.
They come with risks of distorting the text's original meaning, and specifically, the random insertion and random swap tend to break the original structure of the sentence.
These make the text encoder harder to comprehend the augmented text.
Consequently, it is reasonable that EDA, selecting one among these at random, does not bring performance enhancement.

\paragraph{Together with Optimal Augmentations.}
As shown in Table~\ref{table:aug_all}, based on the CLIP$^*$, after combining all optimal data augmentations (\emph{i.e.}, `Augmentation Pool' for image, `Stacking Together' for text), the Rank-1 accuracy is significantly increased by $2.44\%$.
It is worth stressing that the gain is only from exploiting data augmentations.

\subsection{Ablations of Loss Functions}
Table~\ref{table:loss} studies the effectiveness of the loss functions.
(1) Compared to CLIP$^*$+Aug in which the original image-text contrastive loss is used, replacing the loss with the normalized image-text contrastive loss (N-ITC) brings a slight improvement of $0.13\%$ at Rank-1.
(2) For the loss functions that aim to improve data efficiency, it is shown that the image self-supervision (SS-I) achieves the best performance among SS-I, SS-T and SS-IT, and similarly, the multi-view supervision of image (MVS-I) gets the best within MVS-I, MVS-T and MVS-IT. These indicate that improving data efficiency by exploiting image data is more effective than text data in TBPS.
(3) For the loss functions that are optimized for retrieval, both R-ITC and C-ITC improve the performance. In particular, R-ITC leads to a significant boost of $1.41\%$ than CLIP+Aug, showing the significance of R-ITC with the constraint of pulling negative samples apart.
Finally, we combine all of these effective losses, boosting performance by a large margin of $2.76\%$ at Rank-1 based on CLIP$^*$+Aug, achieving $69.54\%$ Rank-1 accuracy.

\begin{table}[t]
\setlength{\abovecaptionskip}{-0.01cm}
\begin{center}
{\caption{Comparisons with state-of-the-art methods on CUHK-PEDES.}\label{table:cmp_cuhk}}
\resizebox{0.49\textwidth}{!}{
\begin{tabular}{l!{\vrule}l!{\vrule}cccc}
\toprule
Methods & References & Rank-1 & Rank-5 & Rank-10 & mAP   \\
\midrule
\multicolumn{6}{l}{\textit{w/o CLIP:}} \\
\midrule
ViTAA~\cite{wang2020vitaa} &ECCV20& 55.97 & 75.84 & 83.52 & - \\
DSSL~\cite{zhu2021dssl} &MM21& 59.98 & 80.41 & 87.56 & - \\
SSAN~\cite{ding2021semantically} &arXiv21& 61.37 & 80.15 & 86.73 & - \\
LapsCore~\cite{wu2021lapscore} &ICCV21& 63.40 & - & 87.80 & - \\
CAIBC~\cite{wang2022caibc} &MM22& 64.43 & 82.87 & 88.37 & - \\
LGUR~\cite{shao2022learning} &MM22& 65.25 & 83.12 & 89.00 & - \\
AXM-Net~\cite{farooq2022axm} &AAAI22& 64.44 & 80.52 & 86.77 & 58.73 \\
IVT~\cite{shu2023see} &ECCVW22& 65.59 & 83.11 & 89.21 & 60.66 \\
SAF~\cite{li2022learning} &ICASSP22& 64.13 & 82.62 & 88.40 & 58.61 \\
RaSa~\cite{bai2023rasa} &IJCAI23&   \textbf{76.51} &   \textbf{90.29} &   \textbf{94.25} &   \textbf{69.38} \\
\midrule
\multicolumn{6}{l}{\textit{w/ CLIP:}} \\
\midrule
TBPS-LD~\cite{han2021text} &BMVC21& 64.08 & 81.73 & 88.19 & 60.08 \\
CFine~\cite{yan2022clip} &arXiv22& 69.57 & 85.93 & 91.15 & - \\
TP-TPS~\cite{wang2023exploiting} &arXiv23& 70.16 & 86.10 & 90.98 &   66.32 \\
IRRA (ViT-B/16)~\cite{jiang2023cross} &CVPR23& 73.38 &   \textbf{89.93} &   \textbf{93.71} & \textbf{66.13} \\
\midrule
CLIP (ViT-B/32) &Baseline& 60.67 & 81.99 & 88.87 & 54.72 \\
TBPS-CLIP (ViT-B/32) &Ours& 69.54 & 86.99 & 91.24 & 61.57 \\
CLIP (ViT-B/16) &Baseline& 65.37 & 85.83 & 91.59 & 59.42 \\
  TBPS-CLIP (ViT-B/16) &Ours&   \textbf{73.54} & 88.19 & 92.35 & 65.38 \\
\hdashline
Simplified TBPS-CLIP (ViT-B/16) &Ours& 72.66 & 88.14 & 92.72 & 64.97 \\
\bottomrule
\end{tabular}
}
\end{center}
\vspace{-0.1cm}
\end{table}

\begin{table}
\setlength{\abovecaptionskip}{-0.01cm}
\begin{center}
\caption{Comparisons with state-of-the-art methods on ICFG-PEDES.}\label{table:cmp_icfg}
\resizebox{0.49\textwidth}{!}{
\begin{tabular}{l!{\vrule}l!{\vrule}cccc}
\toprule
Methods & References & Rank-1 & Rank-5 & Rank-10 & mAP   \\
\midrule
\multicolumn{6}{l}{\textit{w/o CLIP:}} \\
\midrule
ViTAA~\cite{wang2020vitaa} &ECCV20& 50.98 & 68.79 & 75.78 & - \\
SSAN~\cite{ding2021semantically} &arXiv21& 54.23 & 72.63 & 79.53 & - \\
LGUR~\cite{shao2022learning} &MM22& 57.42 & 74.97 & 81.45 & - \\
IVT~\cite{shu2023see} &ECCVW22& 56.04 & 73.60 & 80.22 & - \\
SAF~\cite{li2022learning} &ICASSP22& 54.86 & 72.13 & 79.13 & 32.76 \\
RaSa~\cite{bai2023rasa} &IJCAI23&   \textbf{65.28} &   \textbf{80.40} &   \textbf{85.12} &   \textbf{41.29} \\
\midrule
\multicolumn{6}{l}{\textit{w/ CLIP:}} \\
\midrule
TP-TPS~\cite{wang2023exploiting} &arXiv23& 60.64 & 75.97 & 81.76 &  \textbf{42.78} \\
CFine~\cite{yan2022clip} &arXiv22& 60.83 & 76.55 & 82.42 & - \\
IRRA (ViT-B/16)~\cite{jiang2023cross} &CVPR23& 63.46 & 80.25 &  \textbf{85.82} & 38.06 \\
\midrule
CLIP (ViT-B/32) &Baseline& 53.96 & 73.69 & 80.43 & 32.37 \\
TBPS-CLIP (ViT-B/32) &Ours& 59.88 & 77.40 & 83.33 & 34.96  \\
CLIP (ViT-B/16) &Baseline& 55.97 & 74.62 & 81.35 & 30.63 \\
  TBPS-CLIP (ViT-B/16) &Ours&   \textbf{65.05} &   \textbf{80.34} & 85.47 & 39.83 \\
\hdashline
Simplified TBPS-CLIP (ViT-B/16) &Ours& 64.52 & 80.03 & 85.39 & 39.54
 \\
\bottomrule
\end{tabular}
}
\end{center}
\vspace{-0.5cm}
\end{table}

\begin{table}[t]
\setlength{\abovecaptionskip}{-0.01cm}
\begin{center}
{\caption{Comparisons with state-of-the-art methods on RSTPReid.}\label{table:cmp_rstp}}
\resizebox{0.49\textwidth}{!}{
\begin{tabular}{l!{\vrule}l!{\vrule}cccc}
\toprule
Methods & References & Rank-1 & Rank-5 & Rank-10 & mAP \\
\midrule
\multicolumn{6}{l}{\textit{w/o CLIP:}} \\
\midrule
DSSL~\cite{zhu2021dssl} &MM21& 39.05 & 62.60 & 73.95 & - \\
SSAN~\cite{ding2021semantically} &arXiv21& 43.50 & 67.80 & 77.15 & - \\
IVT~\cite{shu2023see} &ECCVW22& 46.70 & 70.00 & 78.80 & - \\
SAF~\cite{li2022learning} &ICASSP22 & 44.05 & 67.30 & 76.25 & 36.81 \\
RaSa~\cite{bai2023rasa} &IJCAI23&   \textbf{66.90} &   \textbf{86.50} &   \textbf{91.35} &   \textbf{52.31} \\
\midrule
\multicolumn{6}{l}{\textit{w/ CLIP:}} \\
\midrule
CFine~\cite{yan2022clip} &arXiv22& 50.55 & 72.50 & 81.60 & - \\
TP-TPS~\cite{wang2023exploiting} &arXiv23& 50.65 & 72.45 & 81.20 & 43.11 \\
IRRA (ViT-B/16)~\cite{jiang2023cross} &CVPR23& 60.20 & 81.30 & 88.20 & 47.17 \\
\midrule
CLIP (ViT-B/32) &Baseline& 50.10 & 76.10 & 84.95 & 41.14 \\
TBPS-CLIP (ViT-B/32) &Ours& 56.65 & 80.75 & 87.30 & 44.00 \\
CLIP (ViT-B/16) &Baseline& 56.15 & 78.30 & 86.60 & 43.26 \\
TBPS-CLIP (ViT-B/16) &Ours& 61.95 &   \textbf{83.55} &   \textbf{88.75} &   \textbf{48.26} \\
\hdashline
  Simplified TBPS-CLIP (ViT-B/16) &Ours&   \textbf{62.10} & 81.90 & 87.75 & 48.00
 \\
\bottomrule
\end{tabular}
}
\end{center}
\vspace{-0.1cm}
\end{table}

\subsection{Comparisons with State-of-the-Art Methods} 
We compare TBPS-CLIP with state-of-the-art methods
on CUHK-PEDES, ICFG-PEDES and RSTPReid datasets, as shown in Table~\ref{table:cmp_cuhk}, Table~\ref{table:cmp_icfg} and Table~\ref{table:cmp_rstp}, respectively.

(1) In comparison with the methods using CLIP, the proposed TBPS-CLIP with ViT-B/16 as the image encoder outperforms the state-of-the-art method IRRA by a margin on ICFG-PEDES and RSTPReid, and performs on par with it on CUHK-PEDES.
It is worth noting that IRRA appends a multimodal interaction encoder after CLIP for obtaining superior performance, while the proposed TBPS-CLIP keeps it simple in the network architecture (\emph{i.e.,} original two-stream network architecture of CLIP) and also achieves promising results.
\textbf{It can be clearly seen from Table~\ref{table:cmp_sota} that TBPS-CLIP is more lightweight than IRRA.
TBPS-CLIP prompts CLIP to utilize and mine data efficiently, enabling the network to be trained merely with 5 epochs. 
The highly efficient training makes it very friendly as a baseline.}
(2) In comparison with the methods without CLIP, we notice that RaSa performs strongly.
It adopts ALBEF~\cite{li2021align} as the baseline and contains two models, an online model and its momentum version, each consisting of an image encoder, a text encoder and a cross-modal encoder.
\textbf{Although having outstanding performance, RaSa is cumbersome and challenging to support its generalization, as shown in Table~\ref{table:cmp_sota}.
The proposed TBPS-CLIP, with very lightweight and low-cost architecture and promising performance, has the potential as a baseline to provide broader applications.}
(3) Considering the convenience of applying TBPS-CLIP as the baseline, we further provide a simplified TBPS-CLIP, in which there are only N-ITC and R-ITC losses. It still performs satisfactorily, specifically, even beats IRRA on ICFG-PEDES and RSTPReid.
The simplified TBPS-CLIP equipped with two losses can be more easily applied as a baseline in further work.

\begin{table}[t]
\setlength{\abovecaptionskip}{-0.01cm}
\begin{center}
{\caption{Comparisons with recently SOTA methods on CUHK-PEDES. Param.(M) and Epoch denote the number of modal parameters  (in millions) and the number of epochs in training, respectively. Time(s) represents the online running time (in seconds).}\label{table:cmp_sota}}
\resizebox{0.49\textwidth}{!}{
\begin{tabular}{l!{\vrule}l!{\vrule}cccc}
\toprule
\multirow{2}{*}{Methods} & \multirow{2}{*}{Baselines} & \multirow{2}{*}{Param.(M)} & \multirow{2}{*}{Epoch} & \multicolumn{2}{c}{Time(s)} \\
&  &  &  & Training & Test \\
\midrule
RaSa~\cite{bai2023rasa} & ALBEF & 210.2 & 30 & 27967.5 & 869.8 \\
IRRA~\cite{jiang2023cross} & CLIP (ViT-B/16) & 194.5 & 60 & 6110.4 & 31.4 \\
TBPS-CLIP (Ours) & CLIP (ViT-B/16) & 149.2 & 5 & 1234.7 & 31.4 \\
\bottomrule
\end{tabular}
}
\end{center}
\vspace{-0.5cm}
\end{table}

\begin{table}[t]
\setlength{\abovecaptionskip}{-0.01cm}
\begin{center}
{\caption{Results for applying the proposed TBPS-CLIP as the baseline of other method on CUHK-PEDES. The mean Inverse Negative Penalty(mINP) metric is used in IRRA and also adopted here for comparison.}\label{table:genera}}
\resizebox{0.49\textwidth}{!}{
\begin{tabular}{l!{\vrule}l!{\vrule}ccccc}
\toprule
Methods & Baselines & Rank-1 & Rank-5 & Rank-10 & mAP & mINP \\
\midrule
IRRA & CLIP & 73.38 & 89.93 & 93.71 & 66.13 & 50.24 \\
\hdashline
IRRA$^*$ & TBPS-CLIP & 74.97 & 89.82 & 93.80 & 67.84 & 52.53 \\
IRRA$^*$ & Simplified TBPS-CLIP & 74.56	& 89.26	& 93.52	& 67.52 & 52.58 \\
\bottomrule
\end{tabular}
}
\end{center}
\vspace{-0.1cm}
\end{table}

\begin{figure}[h]
\setlength{\abovecaptionskip}{0.05cm}
    \centering
    \subfloat[TBPS-CLIP]{\includegraphics[width=0.48\columnwidth]{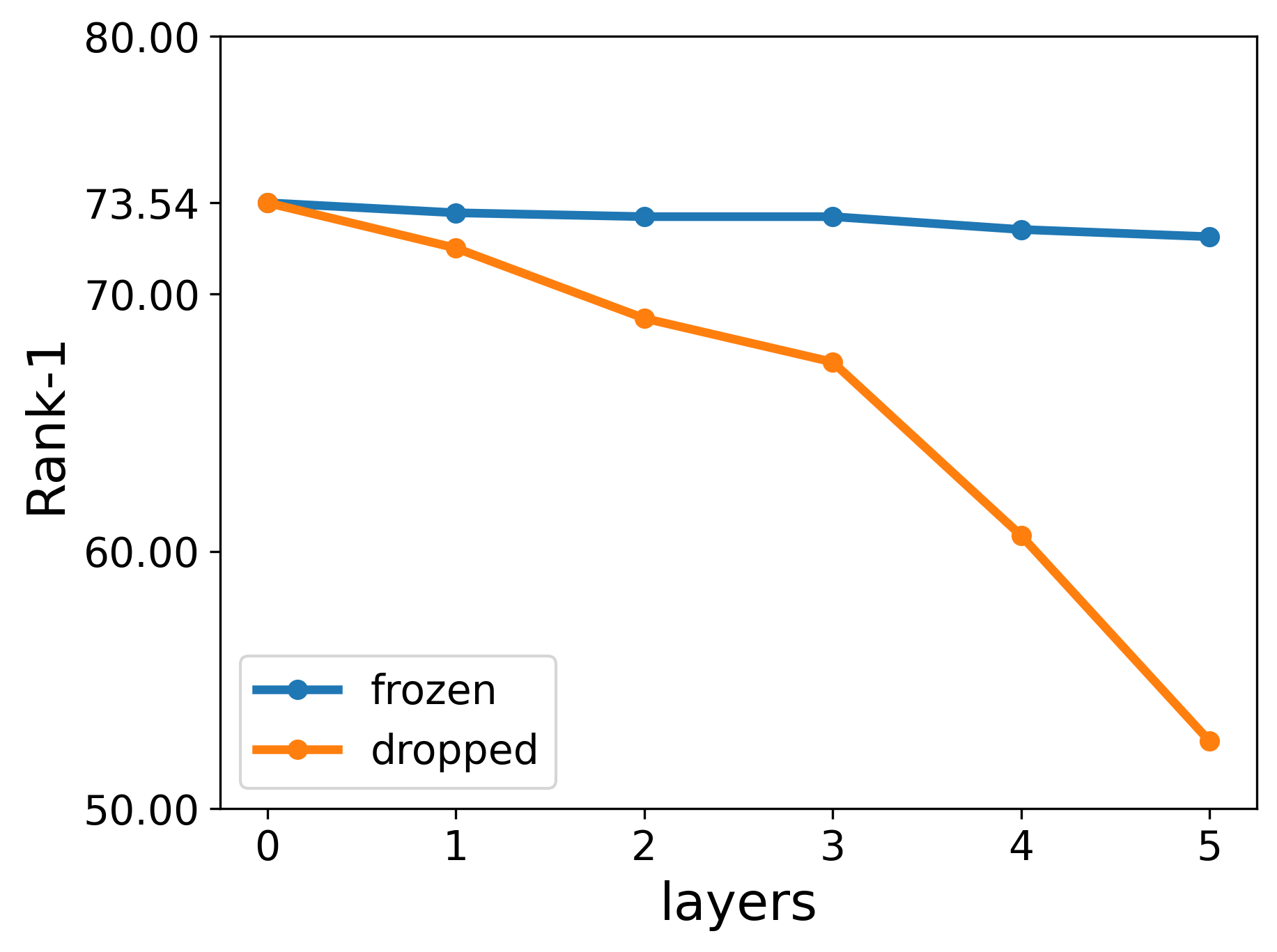}}
    \hspace{0.02\columnwidth}
    \subfloat[Simplified TBPS-CLIP]{\includegraphics[width=0.48\columnwidth]{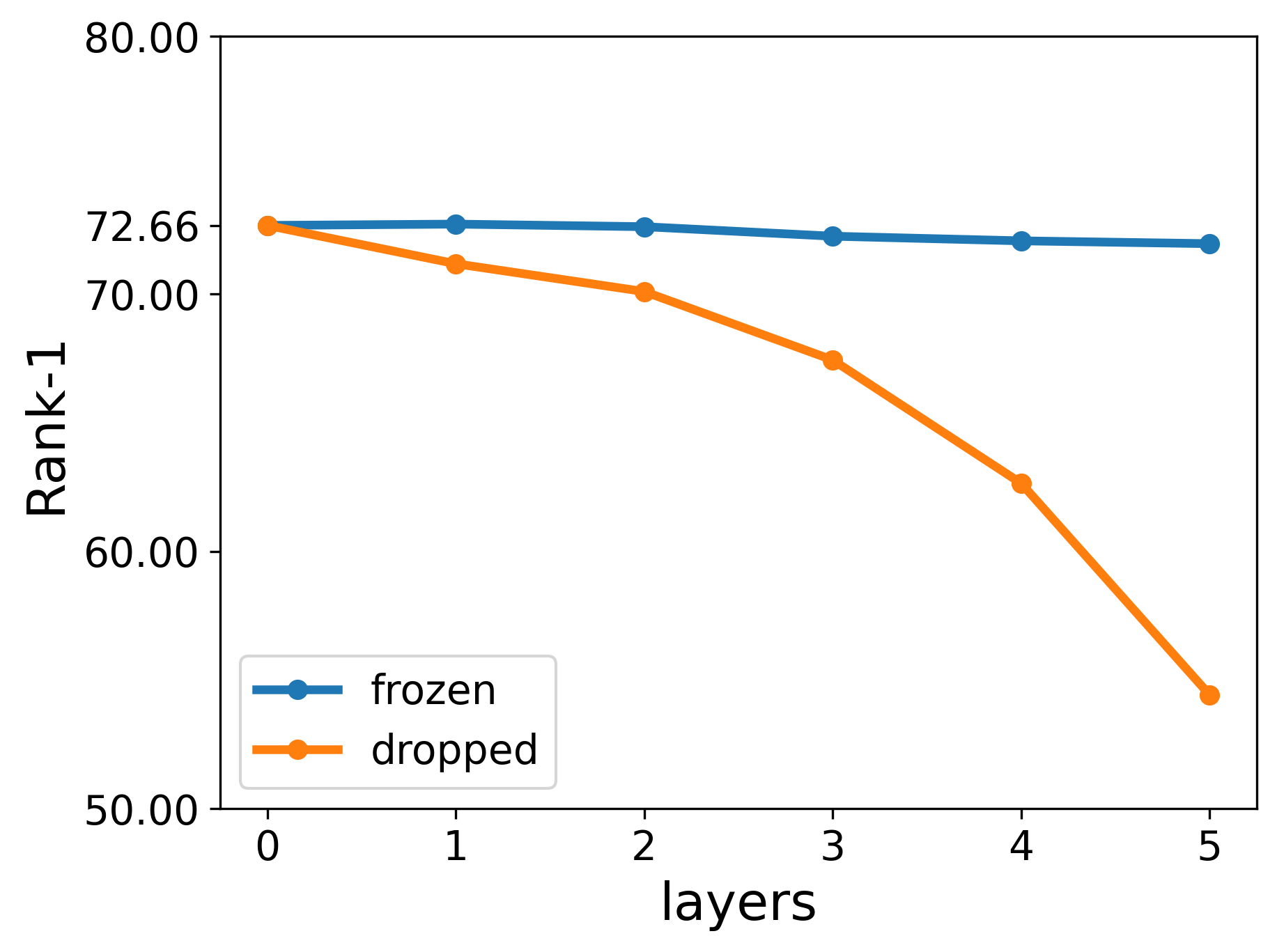}}
    \caption{The trend of performance when dropping/freezing some layers of the text encoder in the proposed TBPS-CLIP and simplified TBPS-CLIP.}
    \label{fig:plotimage}
    \vspace{-0.5cm}
\end{figure}

\begin{table}[t]
\setlength{\abovecaptionskip}{-0.01cm}
\begin{center}
{\caption{Performance under few-shot settings (5\% training data) on CUHK-PEDES.}\label{table:few_shot}}
\resizebox{0.47\textwidth}{!}{
\begin{tabular}{l!{\vrule}cccc}
\toprule
Methods  & Rank-1 & Rank-5 & Rank-10 & mAP \\
\midrule
CLIP~\cite{radford2021learning} & 38.37 & 62.26 & 72.34 & 35.39 \\
CoOp~\cite{zhou2022learning} & 11.37 & 24.19 & 32.46 & 10.48 \\
CLIP-Adapter~\cite{gao2021clip} & 11.96  & 25.45  & 33.56  & 10.91  \\
\hdashline
TBPS-CLIP & 42.98 & 66.26 & 74.94 & 38.86 \\
Simplified TBPS-CLIP & 43.65 & 66.60 & 75.91 & 39.30 \\
\bottomrule
\end{tabular}}
\end{center}
\vspace{-0.5cm}
\end{table}





\subsection{More Probing Experiments of TBPS-CLIP}

\subsubsection{Model Generalization.}
(1) We select IRRA~\cite{jiang2023cross}, the most advanced CLIP-based TBPS method so far, as the hotbed for verifying the generalization and effectiveness of TBPS-CLIP baseline.
Specifically, we adopt TBPS-CLIP (ViT-B/16) and its simplified version as the IRRA's baseline, respectively, instead of the original CLIP (ViT-B/16).
Table~\ref{table:genera} demonstrates the generalization and effectiveness of TBPS-CLIP.
\emph{More experimental results on other datasets are shown in the Appendix.}
(2) We study the few-shot capabilities (5\% training data) of TBPS-CLIP in Table~\ref{table:few_shot}, and \emph{more experimental results with 1\% training data and 10\% one are shown in the Appendix.}
CLIP presents a poor performance in few-shot TBPS, especially, the representative few-shot CLIP variants (CoOp and CLIP-Adapter) are even worse in performance.
The three methods are skilled in the few-shot image classification since the large-scale data knowledge of CLIP from the pre-training phase shares the same data characteristics as the downstream classification task.
However, TBPS is a fine-grained person-specific task and has a noticeable gap with the pre-trained data in CLIP.
The TBPS performance will be poor if there is insufficient training data to fine-tune CLIP,
and CoOp and CLIP-Adapter with the locked CLIP backbone in the fine-tuning phase also perform poorly in TBPS.
Alternatively, the proposed TBPS-CLIP with powerful learning capacity in TBPS can alleviate these problems and brings promising few-shot results.
More than that, the simplified TBPS-CLIP has superiority in the few-shot setting.



\subsubsection{Model Compression.}



We compute the two metrics of TBPS-CLIP in Section~\ref{sec:3.4} as the guidance for the model compression.
\emph{The computation of the metrics, the corresponding investigation of the internal properties of TBPS-CLIP, as well as the detailed explanation of model compression are shown in Appendix.}
Finally, from Figure~\ref{fig:plotimage}, we can clearly see that freezing some layers of the text encoder does not have much impact on performance while the dropping operation can negatively affect performance.
As a result, we can compress TBPS-CLIP during training by freezing part of text encoder.

\section{Conclusion}

This paper makes a thorough empirical study to explore the potential of CLIP for TBPS.
We empirically prove that CLIP, only equipped with some common data augmentations, loss functions, and practical training tricks (without introducing any sophisticated module), can achieve promising results on multiple TBPS benchmarks.
Further, we prove the effectiveness of the proposed TBPS-CLIP from two viewpoints: model generalization and model compression. 
The empirical study is expected to provide a practical guideline for future CLIP-based TBPS research.


\bibliography{aaai24}


\appendix

\section{Empirical Studies}

\subsection{Training Tricks}
\textbf{Global Gradients Back-propagation.}
In multi-GPU training scenarios, data parallelism is often employed to increase the batch size and thus enhance performance.
In such a case, a default practice of code implementation is to only back-propagate the gradient of the loss functions from a local GPU while discarding the gradients gathered from other GPUs. 
We experimentally observe a notable performance gain when back-propagating all gradients across GPUs.

\textbf{Dropout.}
As a regularization trick, dropout~\cite{srivastava2014dropout} is commonly employed by randomly setting some of the neurons' outputs to zero during model training.
In this study, we apply it to the self-attention modules of the text encoder empirically with a probability of $0.05$. 

\textbf{Locking Bottom Layers.}
When fine-tuning the model, a practical technique is to freeze a portion of the lower layers of the network or to assign them a reduced learning rate. In this study, we observe a slightly better performance when the first convolutional layer in the image encoder, which maps image patches to tokens, is locked.

\textbf{Soft Label.}
To stabilize the training process, we draw inspiration from ALBEF~\cite{li2021align} and apply the soft label to the loss function.
Specifically, let the pseudo-label represent the matching probability between the image and text, as output from the model. The soft label is then the average value of the pseudo-label and ground-truth label.

\subsection{Data Augmentation}

\begin{figure}
\setlength{\abovecaptionskip}{0.05cm}
\centerline{\includegraphics[width=0.98\columnwidth,height=0.52\linewidth]{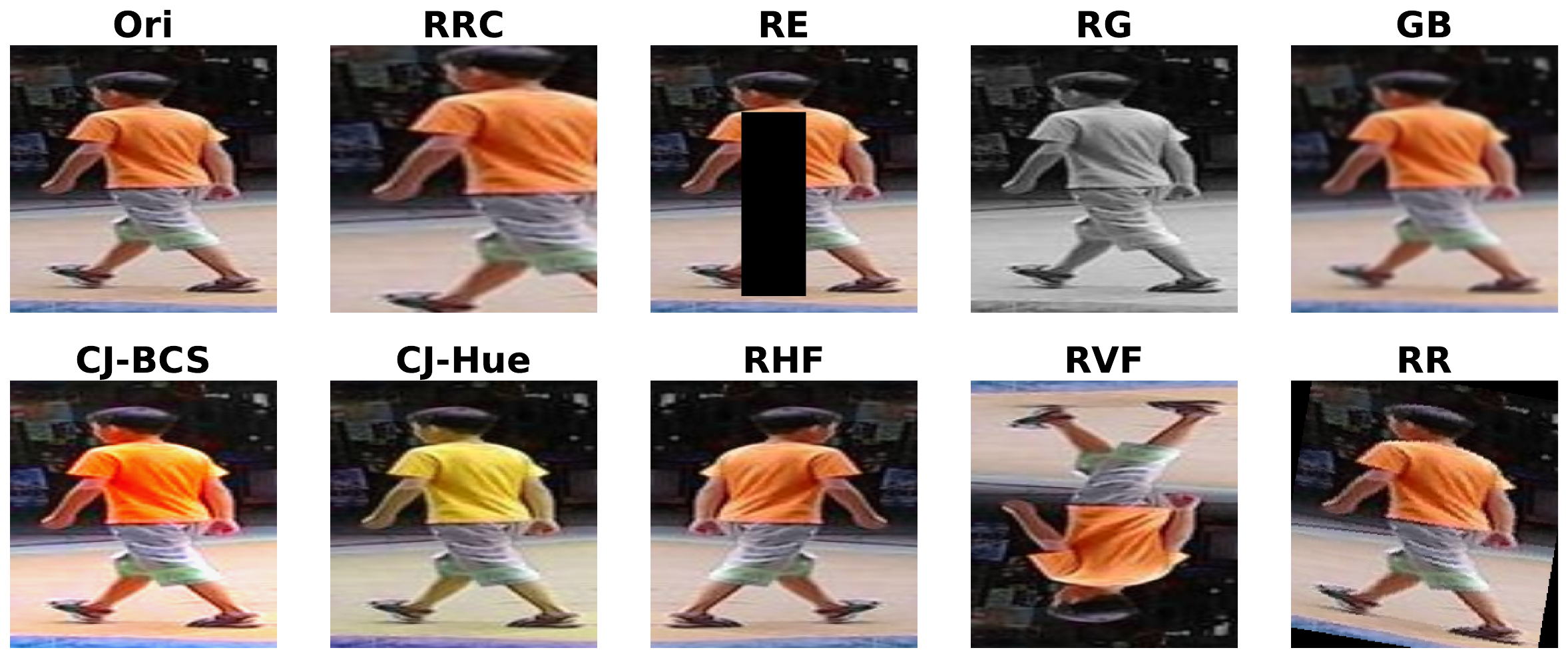}}
\caption{Illustration of the image and its augmented counterparts.
\textbf{Ori:} the original image. 
\textbf{RRC:} RandomResizedCrop. 
\textbf{RE:} RandomErasing. 
\textbf{RG:} RandomGrayscale. 
\textbf{GB:} GaussianBlur. 
\textbf{CJ-BCS:} ColorJitter with adjusted brightness, contrast and saturation. 
\textbf{CJ-Hue:} ColorJitter with adjusted hue. 
\textbf{RHF:} RandomHorizontalFlip. 
\textbf{RVF:} RandomVerticalFlip. 
\textbf{RR:} RandomRotation.} \label{fig:img-aug}
\vspace{-0.5cm}
\end{figure}

\paragraph{Image Augmentation.}
The image augmentations are divided into removal and alteration groups, and their illustrations are presented in Figure~\ref{fig:img-aug}.

The group of removal has:
\begin{itemize}
    \item \textit{RandomResizedCrop} crops a random portion of an image, and then resizes it to a pre-set size as a new augmented image. 
    

    \item \textit{RandomErasing} erases a randomly selected rectangle within the image with a probability of $0.5$. 

    \item \textit{RandomGrayscale} converts the inputted RGB image to a grayscale one with a probability of $0.1$, which can be interpreted as removing all the color information from the image.

    \item \textit{GaussianBlur} applies a Gaussian blur filter to the image, equal to removing its finer details while preserving the coarse content.
    
\end{itemize}

The group of alternation has:
\begin{itemize}
    \item \textit{ColorJitter} randomly adjusts the image's brightness, contrast, saturation, and hue. The brightness, contrast and saturation adjustments do not affect the image color, while the hue adjustment does. Thus, we have the former three grouped and studied, denoted as ColorJitter-BCS, while studying the hue separately, denoted as ColorJitter-Hue.
    
    \item \textit{RandomHorizontalFlip} randomly flips the image horizontally with a probability of $0.5$.

    \item \textit{RandomVerticalFlip} randomly flips the image vertically with a probability of $0.5$.

    \item \textit{RandomRotation} rotates the image by an randomly angle.
\end{itemize}

\paragraph{Text Augmentation.}
We illustrate text augmentation in Table~\ref{table:txt-aug-demo}.

\begin{table}
\setlength{\abovecaptionskip}{-0.01cm}
\begin{center}
{\caption{Illustration of the text and its augmented texts.
\textbf{BT:} back translation.
\textbf{SR:} synonym replacement.
\textbf{RI:} random insertion.
\textbf{RS:} random swap.
\textbf{RD:} random deletion.}\label{table:txt-aug-demo}}
\resizebox{\columnwidth}{!}{
\begin{tabular}{c!{\vrule}p{6.0cm}}
\toprule
Augmentations & Sentence \\
\midrule
None & a man with short black hair is wearing a white t-shirt, long black pants, red shoes and has a black book.\\
\midrule
BT & A man with a short black hair is wearing a white T-shirt, long black pants, red shoes, and black books.\\
\midrule
SR & a man with \textbf{brusk} black hair is wearing a white t-shirt, long black pants, red shoes and has a black \textbf{playscript}.\\
\midrule
RI & a man with short black hair is wearing a \textbf{weary} white t-shirt, long black pants, \textbf{fag out} red shoes and has a black book.\\
\midrule
RS & a \textbf{long} with short black hair is wearing a white t-shirt, \textbf{man} black \textbf{black} red shoes and has a \textbf{pants}, book.\\
\midrule
RD & a man with short black is wearing a \textbf{white} t-shirt, long black pants, red shoes and has a black book.\\
\bottomrule
\end{tabular}
}
\end{center}
\vspace{-0.5cm}
\end{table}


\subsection{Model Generalization}
We conduct the few-shot experiments to verify the model generalization.
Among them, two representative few-shot CLIP methods, CoOp~\cite{zhou2022learning} and CLIP-Adapter~\cite{gao2021clip}, are added for the performance comparison. 


CoOp is a prompt-based method. It employs learnable vectors as soft text prompts, which are acquired by training small amounts of samples.
The original CLIP backbone is locked during training.
Specifically, the $M$ learnable tokens are prepended to the text, such that the input of the text encoder can be represented as:
\begin{equation}
    \text{Text} = [\text{V}]_1 [\text{V}]_2 \hdots [\text{V}]_M [\text{T}]_1 [\text{T}]_2 \hdots [\text{T}]_N,
\end{equation}
where $[\text{V}]_m$ ($m=1, \cdots, M$) is a learnable prompt vector and $[\text{T}]_n$ ($n=1, \cdots, N$) is the vector of the original text token.
The learnable prompt is experimentally initialized as `a pedestrian photo of'.


CLIP-Adapter is an adapter-based method, in which a two-layer MLP $\mathcal{G}$ is appended on the image encoder of CLIP in a residual connection manner.
The parameters of $\mathcal{G}$ are trained based on small amounts of data, while the original backbone is locked.
The final representation $\hat{f}$ can be formulated as:
\begin{equation}
    \hat{f} = \alpha \mathcal{G}(f) + (1 - \alpha) f,
\end{equation}
where $\alpha$ is a residual factor and is experimentally set to $0.4$, and $f$ is the representation from the original model.

\subsection{Model Compression}
\label{MMC}
Two evaluation metrics~\cite{wang2020rethinking} are elaborated in detail here.
The first one estimates the contribution of a specific module by removing the module and observing the consequent performance drop.
Formally, let $\Delta{M}_i$ denotes the performance drop when the module $i$ is removed, and the contribution score of the module $i$ is defined as:
\begin{equation}
    C^{\mathrm{1}}_i = \frac{\Delta{M}_i}{\max\{\Delta{M}_1, \dots, \Delta{M}_n\}},
\end{equation}
where $n$ is the number of modules to be evaluated.
A higher $C^{\mathrm{1}}_i$ signifies a greater performance drop when the module is absent, indicating more contribution of this module to the final performance.

The second metric probes the module's importance by measuring how closely the module's weights can close their initial values while maintaining a certain level of performance. Specifically, the contribution score of the module $i$ is defined as:
\begin{equation}
     C^{\mathrm{2}}_i = \min \alpha_i ~~~~  s.t. ~ M(\theta_i)-M(\theta_i^{\alpha_i}) < \epsilon,
     \label{eq:C2}
\end{equation}
where $\theta_i$ is the final optimized weights of the module $i$ and its corresponding model's performance is $M(\theta_i)$, and $\theta_i^{\alpha_i}= (1-\alpha_i)\theta_i^0 + \alpha_i \theta_i$ ($\alpha_i \in [0,1]$) represents a convex combination of the initial weights $\theta_i^0$ and final weights $\theta_i$ and its corresponding model's performance is $M(\theta_i^{\alpha_i})$.


A smaller $\alpha_i$ indicates that $\theta_i^{\alpha_i}$ is closer to the initial weights $\theta_i^0$.
Thus, a smaller $C^{\mathrm{2}}_i$ implies that the weights of the module $i$ can be moved from their final weights to randomly initialized ones, suggesting a lower contribution of the module to the model performance.

\begin{table}[t]
\setlength{\abovecaptionskip}{-0.01cm}
\begin{center}
{\caption{Ablation study of the hyperparameter on image augmentation on CUHK-PEDES. The underlined hyperparameter value is eventually used in the experiments.}\label{table:app-img-aug}}
\resizebox{\columnwidth}{!}{
\begin{tabular}{c!{\vrule}cc!{\vrule}ccc}
\toprule
Method & Hyperparameter & Value & Rank-1 & Rank-5 & mAP \\
\midrule
\multirow{3}{*}{\shortstack{Random\\Resized\\Crop}} & \multirow{3}{*}{$x_{RRC}$} & $0.3$ & 64.82 & 84.94 & 58.33 \\
& & $0.6$ & 65.11 & 84.65 & 58.66 \\
& & \underline{${0.9}$} &  \underline{65.29} &  \underline{85.01} &  \underline{58.62} \\
\midrule
\multirow{3}{*}{\shortstack{Random\\Erasing}} & \multirow{3}{*}{$\left[x_{RE},y_{RE}\right]$} & $\left[0.02, 0.10\right]$ & 64.51 & 84.86 & 58.00 \\
& & $\left[{\underline{0.10}}, {\underline{0.20}}\right]$ &  \underline{64.64} &  \underline{84.34} &  \underline{58.08} \\
& & $\left[0.20, 0.30\right]$ & 64.30 & 84.60 & 57.97 \\
\midrule
\multirow{3}{*}{\shortstack{Gaussian\\Blur}} & \multirow{3}{*}{$k$} & $\underline{3}$ & \underline{55.30} & \underline{77.86} & \underline{49.82} \\
& & $7$ & 47.01 & 71.02 & 42.66 \\
& & $11$ & 46.48 & 70.60 & 42.09 \\
\midrule

\multirow{6}{*}{\shortstack{Color\\Jitter}} & \multirow{3}{*}{$x_{cj1}$} & ${\underline{0.1}}$ &  \underline{64.86} & \underline{84.65} & \underline{57.85} \\
& & $0.2$ & 64.72 & 84.24 & 57.50 \\
& & $0.4$ & 64.23 & 84.73 & 57.52 \\
\cmidrule{2-6}
 & \multirow{3}{*}{$x_{cj2}$} & $\underline{0.1}$ & \underline{64.12} & \underline{84.81} & \underline{57.41}\\
&&$0.2$ & 63.86 & 84.02 & 56.66\\
&&$0.4$ & 63.69 & 84.11 & 57.08\\
\midrule
\multirow{3}{*}{\shortstack{Random\\Rotation}} & \multirow{3}{*}{$degrees$} & $7.5$ & 65.42 & 85.53 & 58.57 \\
& & ${\underline{15}}$ &  \underline{65.50} &  \underline{85.20} &  \underline{58.92} \\
& & $30$ & 65.06 & 84.99 & 58.45 \\
\bottomrule
\end{tabular}}
\end{center}
\vspace{-0.5cm}
\end{table}

\section{Experiments}

\paragraph{Datasets.}
\textbf{CUHK-PEDES}~\cite{li2017person} is the first TBPS dataset published in $2017$ and the most widely used. This dataset has an average of $3$ images per person, and $2$ textual descriptions accompany each image. They are divided into a train set with $34,054$ images of $11,003$ identities, a validation and a test set containing $3,078$ and $3,074$ images of $1,000$ identities, respectively.
\textbf{ICFG-PEDES}~\cite{ding2021semantically} is a larger dataset that contains a train set with $34,674$ image-text pairs of $3,102$ identities and a test set with $19,848$ image-text pairs of $1,000$ identities. The images include more complex backgrounds and illumination, and the text description is longer.
\textbf{RSTPReid}~\cite{zhu2021dssl} is a newly published dataset for suitably handling real scenarios. Each identity has $5$ images captured by different cameras, and each image corresponds to $2$ textual descriptions in this dataset. All are split into three training, validation and test subsets with $3,701$, $200$ and $200$ identities, respectively.

\paragraph{Evaluation Metrics.}
The commonly used Rank-$k$ metric and mean Average Precision (mAP) are adopted as the metrics to evaluate the performance in our experiments. Rank-$k$ demonstrates that at least one correct match is retrieved among results within top $k$ similarities and $k$ is set to $1$, $5$ and $10$ following the common practice. 

\paragraph{Implementation Details.} 
All experiments are conducted on 4 Nvidia A40 GPUs.
The $224 \times 224$ image is divided into $49$ patches with a size of $32 \times 32$ and are inputted to the model (ViT-B/32), and is divided with $16 \times 16$ in the model (ViT-B/16).
The texts with a fixed length of $77$ are inputted to the model.
The AdamW optimizer is employed with a linear warmup and cosine decay schedule, with $1e-6$, $5e-6$ and $1e-4$ for the initial, final and peak learning rate, respectively.
The model is trained for merely 5 epochs.

\subsection{Ablations of Data Augmentation}
The hyperparameters of data augmentation are described as follows.

\subsubsection{Image Augmentation}

\begin{itemize}
    \item \textit{RandomResizedCrop.} For the cropping operation, the aspect ratio of the cropped area is randomly selected in $\left[ \frac{3}{4}, \frac{4}{3} \right]$ (default setting on Pytorch), and the scale of the cropped area with respect to the original image is randomly chosen in $\left[ x_{RRC}, 1 \right]$. The augmented image is then resized to $224 \times 224$. 

    \item \textit{RandomErasing.} The aspect ratio of the erased area is randomly chosen in $\left[ \frac{3}{10}, \frac{10}{3} \right]$. The proportion of the erased area against the original image is selected randomly within the range $\left[ x_{RE}, y_{RE} \right]$. 

    \item \textit{GaussianBlur.} The size of Gaussian kernel is set to $k$, while the standard deviation is randomly chosen between $\left[ \frac{1}{10}, 2 \right]$.

    \item \textit{ColorJitter.}
    The factor of bright brightness, contrast and saturation is chosen uniformly from $\left[ 1-x_{cj1}, 1+x_{cj1} \right]$, while the factor of hue is chosen uniformly from $\left[ -x_{cj2}, +x_{cj2} \right]$. 

    \item \textit{RandomRotation.} The rotation degree is randomly chosen between $\left[ -degree, +degree \right]$.
\end{itemize}

We show the results of applying image augmentation with different hyperparameters mentioned above in Table~\ref{table:app-img-aug}.

\begin{table}[t]
\setlength{\abovecaptionskip}{-0.01cm}
\begin{center}
{\caption{Ablation study of the hyperparameter on text augmentation on CUHK-PEDES. The underlined hyperparameter value is eventually used in the experiments.}\label{table:app-txt-aug}}
\resizebox{0.45\textwidth}{!}{
\begin{tabular}{c!{\vrule}c!{\vrule}cccc}
\toprule
Method & $\alpha$ & Rank-1 & Rank-5 & Rank-10 & mAP \\
\midrule
\multirow{3}{*}{\shortstack{Back\\Translation}} & 0.05 & 64.31 & 84.39 & 90.48 & 57.43 \\
& \underline{ 0.1} & \underline{ 64.77} &  \underline{84.67} & \underline{ 90.56} &  \underline{57.49} \\
& 0.2 & 64.47 & 84.58 & 90.64 & 57.27 \\
\midrule
\multirow{3}{*}{\shortstack{Synonym\\Replacement}} & \underline{0.05} & \underline{63.95} & \underline{83.76} & \underline{89.96} & \underline{56.90} \\
& 0.1 & 61.45 & 81.47 & 88.16 & 54.98 \\
& 0.2 & 59.70 & 81.16 & 87.61 & 53.66 \\
\midrule
\multirow{3}{*}{\shortstack{Random\\Insertion}} & \underline{0.05} & \underline{63.71} & \underline{84.29} & \underline{90.03} & \underline{56.77} \\
& 0.1 & 62.83 & 83.17 & 89.49 & 56.06 \\
& 0.2 & 60.09 & 82.44 & 88.84 & 54.17 \\
\midrule
\multirow{3}{*}{\shortstack{Random\\Swap}} & \underline{0.05} & \underline{56.41} & \underline{79.08} & \underline{86.14} & \underline{49.29} \\
& 0.1 & 51.06 & 74.72 & 83.16 & 45.16 \\
& 0.2 & 49.51 & 73.31 & 81.95 & 43.74 \\
\midrule
\multirow{3}{*}{\shortstack{Random\\Deletion}} &  \underline{0.05} & \underline{65.53} & \underline{84.70} & \underline{90.87} & \underline{57.90} \\
& 0.1 & 64.99 & 85.19 & 90.37 & 57.33 \\
& 0.2 & 64.31 & 83.43 & 90.09 & 56.65 \\
\midrule
\multirow{3}{*}{\shortstack{EDA}} & \underline{0.05} & \underline{63.43} & \underline{83.77} & \underline{90.27} & \underline{56.15} \\
& 0.1 & 62.17 & 82.47 & 88.84 & 54.40 \\
& 0.2 & 60.66 & 81.45 & 88.27 & 53.50 \\
\bottomrule
\end{tabular}
}
\end{center}
\vspace{-0.5cm}
\end{table}

\subsubsection{Text Augmentation}
Following EDA~\cite{wei2019eda}, for \emph{synonym replacement}, \emph{random insertion} and \emph{random swap}, the process is repeated $\alpha L$ times in a sentence, where $L$ denotes the sentence length and $\alpha$ is a hyperparameter indicating the proportion of word modification. \emph{Random deletion} is applied with a probability of $\alpha$.
When applying \emph{back translation}, each sentence is replaced by the back translated one with a probability of $\alpha$.

The influence of the hyperparameter on performance is shown in Table~\ref{table:app-txt-aug}.




\subsection{More Probing Experiments of TBPS-CLIP}

\subsubsection{Model Generalization.}
(1) We adopt the proposed TBPS-CLIP (ViT-B/16) and its simplified version as the IRRA's baseline, respectively.
Table~\ref{table:genera} shows the results on ICFG-PEDES and RST-PReid, which strongly verify the generalization and effectiveness of TBPS-CLIP.
(2) Table~\ref{table:few_shot} shows the few-shot results of several methods, including 1\% training data and 10\% training data, respectively.
The proposed TBPS-CLIP presents the absolute advantage under various few-shot settings.
Specifically, the simplified version is more competitive in performance.

\begin{table}
\setlength{\abovecaptionskip}{-0.01cm}
\begin{center}
{\caption{Results for applying the proposed TBPS-CLIP as the baseline of other method. The mean Inverse Negative Penalty(mINP) metric is used in IRRA and also adopted here for comparison.}\label{table:genera}}
\resizebox{0.49\textwidth}{!}{
\begin{tabular}{l!{\vrule}l!{\vrule}ccccc}
\toprule
Methods & Baselines & Rank-1 & Rank-5 & Rank-10 & mAP & mINP \\
\midrule
\multicolumn{7}{l}{\textit{ICFG-PEDES}} \\
\midrule
IRRA & CLIP & 63.46	& 80.25	& 85.82	& 38.06	& 7.93 \\
IRRA$^*$ & TBPS-CLIP & 65.64 & 81.25 & 86.61 & 40.77 & 9.36 \\
IRRA$^*$ & Simplified TBPS-CLIP & 64.90 & 80.95 & 86.48 & 40.50 & 9.40 \\
\midrule
\multicolumn{7}{l}{\textit{RST-PReid}} \\
\midrule
IRRA & CLIP & 60.20 & 81.30 & 88.20 & 47.17 & 25.28 \\
IRRA$^*$ & TBPS-CLIP & 63.40 & 82.60 & 89.00 & 49.45 & 26.78 \\
IRRA$^*$ & Simplified TBPS-CLIP & 62.35 & 82.00 & 87.90 & 49.82 & 27.41 \\
\bottomrule
\end{tabular}
}
\end{center}
\vspace{-0.1cm}
\end{table}

\begin{table}
\setlength{\abovecaptionskip}{-0.01cm}
\begin{center}
{\caption{Performance under few-shot settings on CUHK-PEDES.}\label{table:few_shot}}
\resizebox{0.47\textwidth}{!}{
\begin{tabular}{l!{\vrule}cccc}
\toprule
Methods  & Rank-1 & Rank-5 & Rank-10 & mAP \\
\midrule
\multicolumn{5}{l}{\textit{1\% training data:}} \\
\midrule
CLIP~\cite{radford2021learning} & 27.84 & 49.51 & 60.56 & 25.88 \\
CoOp~\cite{zhou2022learning} & 10.98 & 23.98 & 31.50 & 10.22 \\
CLIP-Adapter~\cite{gao2021clip} & 11.22 & 24.22 & 31.71 & 10.30 \\
\hdashline
TBPS-CLIP & 27.94 & 51.20 & 62.07 & 27.60 \\
Simplified TBPS-CLIP & 30.05 & 51.56 & 62.23 & 28.53 \\
\midrule
\multicolumn{5}{l}{\textit{10\% training data:}} \\
\midrule
CLIP~\cite{radford2021learning} & 42.37 & 66.13 & 75.31 & 38.20  \\
CoOp~\cite{zhou2022learning} &  12.10 & 25.55 & 33.97 & 10.96  \\
CLIP-Adapter~\cite{gao2021clip} & 12.61 & 27.19 & 35.32 & 11.77 \\
\hdashline
TBPS-CLIP & 49.19 & 71.98 & 80.46 & 43.81 \\
Simplified TBPS-CLIP & 50.26 & 72.51 & 80.28 & 44.28 \\
\bottomrule
\end{tabular}}
\end{center}
\vspace{-0.5cm}
\end{table}

\begin{table}[t]
\setlength{\abovecaptionskip}{-0.01cm}
\begin{center}
{\caption{The ID of the dropped/frozen layers in the text encoder, and the corresponding number of the trainable parameters.}\label{table:layer-ID}}
\resizebox{0.43\textwidth}{!}{
\begin{tabular}{l!{\vrule}cccc}
\toprule
\multirow{2}{*}{Layers} & \multirow{2}{*}{ID} & \multicolumn{2}{c}{Trainable Parameters} \\
 & & TBPS-CLIP & Simplified TBPS-CLIP \\
\midrule
0 & - & 149.2 & 149.0 \\
1 & 7 & 146.0 ($\downarrow{2\%}$) & 145.9 ($\downarrow{2\%}$) \\
2 & 7,8 & 142.9 ($\downarrow{4\%}$) & 142.7 ($\downarrow{4\%}$) \\
3 & 3,7,8 & 139.7 ($\downarrow{6\%}$) & 139.6 ($\downarrow{6\%}$)\\
4 & 3,5,7,8 & 136.6 ($\downarrow{8\%}$) & 136.4 ($\downarrow{8\%}$)\\
5 & 3,5,6,7,8 & 133.4 ($\downarrow{11\%}$) & 133.3 ($\downarrow{11\%}$)\\
\bottomrule
\end{tabular}
}
\end{center}
\vspace{-0.1cm}
\end{table}

\subsubsection{Model Compression.}

\begin{figure}[t]
\setlength{\abovecaptionskip}{0.08cm}
    \centering
    \subfloat[Txt enc. $C^\mathrm{1}_i$]{\includegraphics[width=0.25\columnwidth,height=0.36\textwidth]{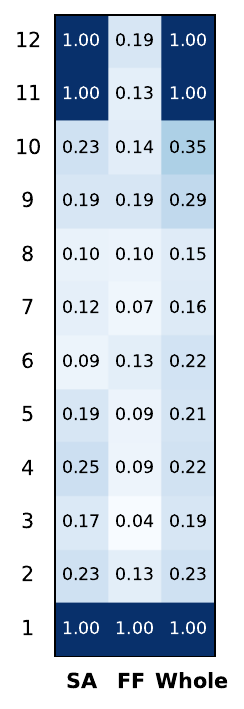}}
    \subfloat[Txt enc. $C^{\mathrm{2}}_i$]{\includegraphics[width=0.25\columnwidth,height=0.36\textwidth]{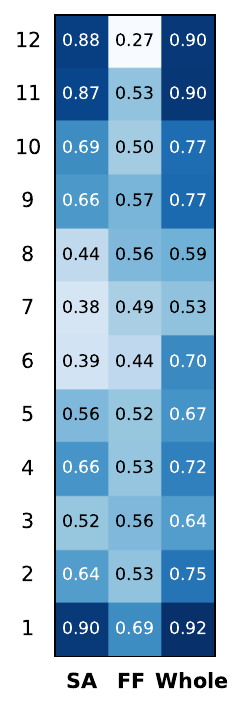}}
    \subfloat[Img enc. $C^\mathrm{1}_i$]{\includegraphics[width=0.25\columnwidth,height=0.36\textwidth]{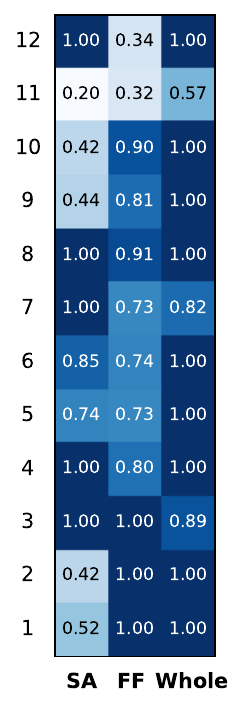}}
    \subfloat[Img enc. $C^{\mathrm{2}}_i$]{\includegraphics[width=0.25\columnwidth,height=0.36\textwidth]{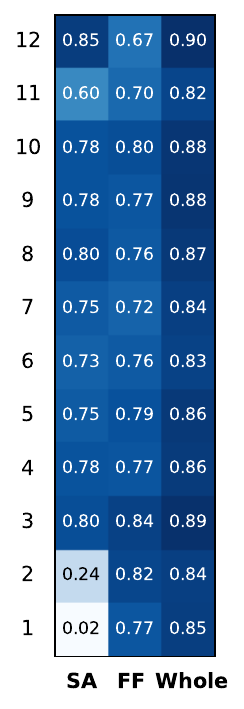}}
    \caption{Contributions of each module in the image encoder (Img enc.) and text encoder (Txt enc.) of TBPS-CLIP on CUHK-PEDES. SA: self-attention module. FF: feed-forward module. Whole: a whole layer in transformer.}
    \label{fig:importance}
    \vspace{-0.5cm}
\end{figure}

The proposed TBPS-CLIP has the same architecture as CLIP, in which there are a text encoder and an image one, both with 12 layers of transformer and each layer consists of one self-attention module and one feed-forward module.
Although it is already a lightweight architecture compared to other TBPS methods, we also consider its model compression for more convenient application in the industry.

We first explore the internal properties and functionalities of TBPS-CLIP by computing the contribution score of each module with the aid of two metrics $C^{\mathrm{1}}_i$ and $C^{\mathrm{2}}_i$ in Section~\ref{MMC}, in which we select the Rank-1 as the performance assessment and set $\epsilon$ to $3$ in the metric $C^{\mathrm{2}}_i$.

As shown in Figure~\ref{fig:importance}, (1)  the results of the two metrics are in good agreement with each other and reveal a common pattern; (2) for the text encoder, the top and bottom layers matter the most, indicating redundancy in the middle layers, and the self-attention module shows more importance than the feed-forward module in each layer; (3) for the image encoder, the contributions of different layers are relatively evenly distributed, and specifically the self-attention modules of the first two layer have less contribution to the final performance.
It could be concluded that the text encoder has architecture redundancy, and removing middle layers maybe have little influence over performance, while the image encoder plays an essential role in achieving outstanding TBPS performance. 

These findings guide the model compression.
We drop or freeze some layers of the text encoder during training.
Specifically, there are $C(12,x)$ combinations to drop/freeze $x$ layers among $12$-layers of text encoder, and we calculate the sum of $x$ layers' Whole scores $C^1$ and $C^2$ in Figure~\ref{fig:importance}, by which the combination with the minimum value is adopted.
We conduct the experiments with $x=1,2, \cdots, 5$.
Referring to Table~\ref{table:layer-ID} for details on which layers are dropped/frozen and the corresponding number of trainable parameters.

\section{Further Discussions}

We summarize some empirical observations as follows.
\begin{enumerate}[label=\arabic*), ref=\arabic*]
\item The CLIP with four training tricks yields about $4\%$ improvement at Rank-1 in Table 1 of the main paper. It can inspire future works in which the model performance could be boosted by applying these training tricks.
\item Data augmentation and loss function are common technologies used in various methods. The investigation of more than $20$ data augmentations and about $10$ loss functions on performance in Tables 2-5 of the main paper provides valuable guidance on future works.
Researchers can select proper and effective augmentations and losses into the model for improving performance.
\item We explore the internal properties and functionalities of the model for the first time. These results can light future works on model compression, so as to develop a more lightweight and effective TBPS method.
\item There are very little research on few-shot TBPS, while this paper makes a preliminary study on CLIP-based few-shot TBPS, providing valuable observation for future research direction. 
\end{enumerate}

\end{document}